\PassOptionsToPackage{dvipsnames,table}{xcolor}
\documentclass[sigconf=true, nonacm=true, review=false, anonymous = false]{acmart}
\usepackage{booktabs} 
   \usepackage[flushleft]{threeparttable}
\usepackage{multirow}
\usepackage{amsmath}
\usepackage{subcaption}
\usepackage{comment}
\usepackage{xcolor}

\usepackage{setspace}

\usepackage[]{caption}

\begin{document}
\title[Comparing AS Approaches on BBO Problems]{Comparing Algorithm Selection Approaches on Black-Box Optimization Problems}

\author{Ana Kostovska}
\orcid{}
\affiliation{%
  \institution{Jo\v{z}ef Stefan Institute}
  \city{Ljubljana} 
 \country{Slovenia}
}

\author{Anja Jankovic}
\orcid{}
\affiliation{%
  \institution{Sorbonne Universit\'e, LIP6}
  \streetaddress{}
  \city{Paris} 
  \country{France}}

\author{Diederick Vermetten}
\orcid{}
\affiliation{%
  \institution{Leiden University, LIACS}
  \streetaddress{}
  \city{Leiden} 
 \country{The Netherlands}
}

\author{Sa\v{s}o D\v{z}eroski}
\orcid{}
\affiliation{%
  \institution{Jo\v{z}ef Stefan Institute}
  \city{Ljubljana} 
 \country{Slovenia}
}

\author{Tome Eftimov}
\orcid{}
\affiliation{%
  \institution{Jo\v{z}ef Stefan Institute}
  \streetaddress{}
  \city{Ljubljana} 
 \country{Slovenia}
}

\author{Carola Doerr} 
\orcid{0000-0002-4981-3227}
\affiliation{%
  \institution{Sorbonne Universit\'e, CNRS, LIP6}
  \streetaddress{}
  \city{Paris} 
  \country{France}}

\renewcommand{\shortauthors}{A. Kostovska et al.}

\begin{abstract}

Performance complementarity of solvers available to tackle black-box optimization problems gives rise to the important task of algorithm selection (AS). Automated AS approaches can help replace tedious and labor-intensive manual selection, and have already shown promising performance in various optimization domains. Automated AS relies on machine learning (ML) techniques to recommend the best algorithm given the information about the problem instance. Unfortunately, there are no clear guidelines for choosing the most appropriate one from a variety of ML techniques. Tree-based models such as Random Forest or XGBoost have consistently demonstrated outstanding performance for automated AS. Transformers and other tabular deep learning models have also been increasingly applied in this context.

We investigate in this work the impact of the choice of the ML technique on AS performance. We compare four ML models on the task of predicting the best solver for the BBOB problems for 7 different runtime budgets in 2 dimensions. While our results confirm that a per-instance AS has indeed impressive potential, we also show that the particular choice of the ML technique is of much minor importance. 
\end{abstract}

\maketitle

\section{Introduction}
\label{sec:intro}

Black-box problems are challenging to optimize as we do not have direct access to information about problem instances at hand, but instead learn about them by sampling and evaluating solution candidates. Black-box optimization algorithms use precisely this information in order to iteratively steer the search towards the optimum. Many algorithms have been developed for the purpose of solving black-box problems, and they exhibit different performances depending on their structural and behavioral properties. Choosing the best-suited algorithm for a certain problem instance is known as per-instance algorithm selection (PIAS), and it has led to impressive performance gains when algorithms available to choose from exhibit complementary performances~\cite{KerschkeHNT19}. Automated PIAS relies on ML techniques to predict which algorithm to recommend based on the characteristics of the problem instance at hand. 
There are many open challenges in adapting PIAS to black-box settings.
One of them concerns the characterization of black-box problem instances without having any prior information about them.
A classical technique to approximate problem instances by computing statistics from a set of solution candidates and their evaluations is known as \emph{Exploratory Landscape Analysis} (ELA)~\cite{mersmann2011exploratory}.
So-obtained \emph{(landscape) features} are then used to discriminate between instances.

Another prominent open question concerns the choice of the ML technique; it is not evident which approaches are best-suited for applications in the context of black-box optimization, and whether there is a big difference between the techniques at all. Given an ELA-based instance representation as input, \emph{feature-based} ML approaches are commonly employed to learn the mapping between them and algorithm performances or rankings for each of those instances as output. Classical supervised approaches for such prediction tasks used in the literature are \emph{regression} and \emph{classification}, as well as a variant of the latter that is \emph{pairwise classification}. Regression models learn to predict a real-valued metric (e.g., target precision or expected runtime), while classification consists of predicting a label of the category in which the input belongs (i.e., the name of the best algorithm). Pairwise classification allows for selecting the overall ``winner'' solver, following a tournament-like structure by converting a multi-class problem into a series of two-class problems. Among different models used for either regression or (pairwise-) classification tasks, classical \emph{tree-based} models, such as Random Forests and XGBoost, have demonstrated tremendous success in algorithm performance prediction in the optimization domain. Recent advances in ML and deep learning (DL) have shown that certain neural network architectures, such as Transformers, could also be successfully applied to tabular data prediction tasks~\cite{grinsztajn2022treebasedtabular, gorishniy2021revisiting}.
Notable examples include TabPFN~\cite{tabpfn}, and FT-Transformer~\cite{gorishniy2021revisiting}. 
A large variety of available techniques yields a meta-optimization problem of choosing the best-performing AS approach or ML model for applications in black-box optimization.

\textbf{Our contribution.} In this paper, we assess the relevance of selecting the underlying ML technique for a given AS task on a single-objective black-box problem set with a portfolio of inherently different black-box solvers.
Specifically, we compare the predictive power of three AS approaches (regression, classification, and pairwise classification), orthogonal to four ML models: two tree-based (Random Forest and XGBoost) in all three AS modalities, and two DL-based models (TabPFN for classification and FT-Transformer for regression) on the same landscape-aware AS task, for seven different solver runtime budgets and in two dimensionalities. We show that the choice of the ML technique has only a minor impact on the AS performance, as well as on the reliability and robustness of the results, as long as it is a method that demonstrates good performance on tabular data in general.
At the same time, we confirm that PIAS truly is a core ingredient for boosting the performance of black-box solvers, irrelevant of the underlying ML technique; any algorithm selection approach outperforms any static choice of the solver across all settings.
We suggest that research efforts should be instead directed towards (1) the identification of (and possibly involving the design of) suitable training sets and (2) the reduction of the overhead cost for feature extraction. 
The choice of the ML technique, in contrast, seems to be much less relevant for the design of efficient PIAS approaches in black-box optimization. 

\textbf{Availability of data and code.} Following best practices towards replicability and reproducibility, full project data, code and detailed description of the experimental setup, as well as figures for all settings, are available at~\cite{gecco2023data}. 

\section{Experimental Setup}
\label{sec:exp-setup}
\subsection{Data Collection}
\label{ssec:data}

\noindent\textbf{Problem instance portfolio.} 
For the problem instance portfolio, we make use of 24 single-objective, noiseless black-box optimization problems from the BBOB benchmark suite of the COCO environment~\cite{hansen2020coco}, whose objectives are to be minimized. In our study, we consider the first 10 instances of the 24 BBOB functions, in dimensions 5 and 20. This results in two separate problem instance portfolios of size 240, one for each dimension.

\noindent\textbf{Problem landscape data.} For representing the problem landscape data, we make use of the so-called "cheap" ELA features implemented in the R package \texttt{flacco}. We have considered a total of 69 different ELA features, which have been calculated using the Latin hypercube sampling strategy with $1\,000$ sample size on a total of 100 independent repetitions. To represent each problem instance, we calculate the median value for each feature over the 100 repetitions.

\noindent\textbf{Algorithm portfolio.} 
For the algorithm portfolio, we use a diverse set of 11 algorithms. For details on the selected algorithms, we refer the reader to \cite{gecco2023data}. 

To collect performance data for our algorithm portfolio, we use the IOHexperimenter framework~\cite{iohexp}. For each problem instance, we perform 50 independent repetitions, for a budget of $10\,000$ function evaluations for each run.
The complementarity of algorithms is essential for reaching peak AS performance. To this end, we have determined the proportion of problem instances where an algorithm is the optimal choice for each runtime, both for the $5D$ and $20D$ problem instance portfolios. Only the algorithms that exhibit superiority over 5\% of the problem instances are included in the final algorithm portfolio. The final algorithm portfolios for different runtimes and problem dimensionalities are made available in \cite{gecco2023data}.

\subsection{Algorithm Performance Prediction}
\label{ssec:ml}

\noindent\textbf{AS approaches.}
To predict the performance of algorithms, we consider three ML-based AS approaches: regression, classification, and pairwise classification. In all three cases, an ELA feature vector is used to describe each problem instance. In the case of regression, we aim to predict the target precision that each algorithm in the portfolio will achieve on the problem instances, for a fixed runtime and problem dimensionality. We cap the target precision at $10^{-8}$ and perform a $\log_{10}$ transformation of the target variable. For classification, to the ELA vector of each problem instance, we assign a label of the best performing algorithm from our portfolio. This translates to the task of multi-class classification, where the target variable is the algorithm name/label we aim to predict. Note that it can happen that multiple algorithms achieve the same target precision on a given problem instance (usually when they reach a target precision of 0, i.e., the optimum). In such cases, we randomly choose one of them as the best. 
Finally, in pairwise classification, we transform the multi-class classification problem into a set of binary classification problems, where we predict which algorithm performs better out of the two on a given problem instance. For an algorithm portfolio of size $K$, in the case of the classification, we have one multi-class classification problem with K classes. In the case of pairwise classification, however, $K(K-1)/2$ binary classification problems need to be solved to cover all possible pairs of algorithms. We count the number of times an algorithm was predicted to be the best among all pairs, and form our final prediction by selecting the algorithm with the most "wins".

\noindent\textbf{ML models.} For each of the AS approaches, we train different ML models to assess whether the performance of the algorithm selector depends on the underlying model. We use Random Forest and XGBoost for all three AS approaches, as well as TabPFN for (pairwise-) classification and FT-Transformer models for regression.

\noindent\textbf{Hyperparameter tuning and model evaluation.} 
In order to evaluate the performance of our ML models, we employ a nested leave-one-group-out cross-validation (CV) technique, where groups are defined based on the ID of each benchmark problem instance. The setup consists of two stages of CV, with an outer loop splitting the data into training and testing sets and an inner loop determining the optimal hyperparameters for the model. We use the grid search strategy for hyperparameter tuning and determine the best hyperparameters based on the average performance on the holdout folds in the inner CV. The $R^2$ score and $ F1$ score are utilized as performance metrics for regression and (pairwise-)classification, respectively. The hyperparameters selected for tuning and their corresponding search spaces are available at~\cite{gecco2023data}. After identifying optimal hyperparameters, the final model is trained on all the training data and evaluated using the test set from the outer loop.

\subsection{Per-instance Algorithm Selection}
\label{ssec:as}
The AS quality is evaluated by comparing it to two standard baselines: (1) the \emph{virtual best solver} (VBS), which represents the performance of a theoretical perfect algorithm selector that always chooses the true best algorithm for each problem instance, and (2) the \emph{single best solver} (SBS), which is the algorithm with the best average performance among all algorithms in the portfolio. 
The ``VBS-SBS gap'' represents the difference between the VBS and SBS performances and gives an indication of the potential gains from PIAS. 
To evaluate the performance of PIAS, we compute the difference (or \emph{loss}) between the target precision of the selected algorithm $F_{\mathcal{A}}$ and the target precision of the per-instance VBS $F_{\mathcal{A}^{*}}$. 
We do this after taking the logarithm of the achieved target precision:
$
\mathcal{L}(\mathcal{A}, \mathcal{A}^{*}) = \log_{10}(F_{\mathcal{A}}) -  \log_{10}(F_{\mathcal{A}^{*}}).
$
We calculate the losses for each instance separately and investigate their distribution.

\section{Results}
\begin{figure*}
\centering
\begin{subfigure}{0.32\linewidth}
\includegraphics[width=\linewidth]{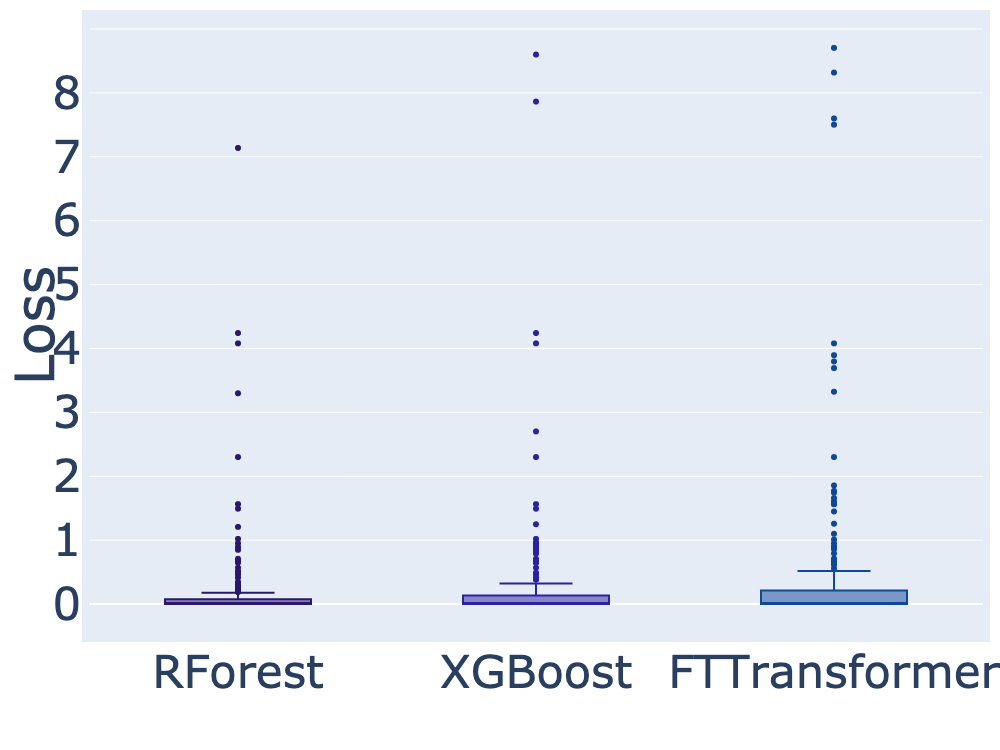}
\caption{regression}
\end{subfigure}
\begin{subfigure}{0.32\linewidth}
\includegraphics[width=\linewidth]{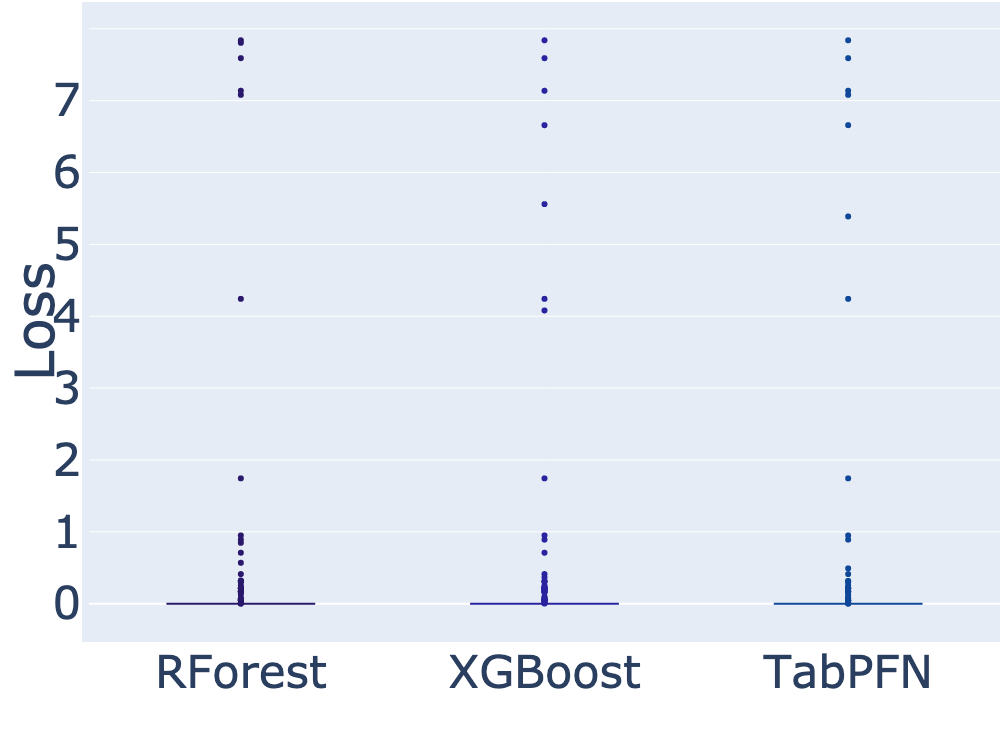}
\caption{classification}
\end{subfigure}
\begin{subfigure}{0.32\linewidth}
\includegraphics[width=\linewidth]{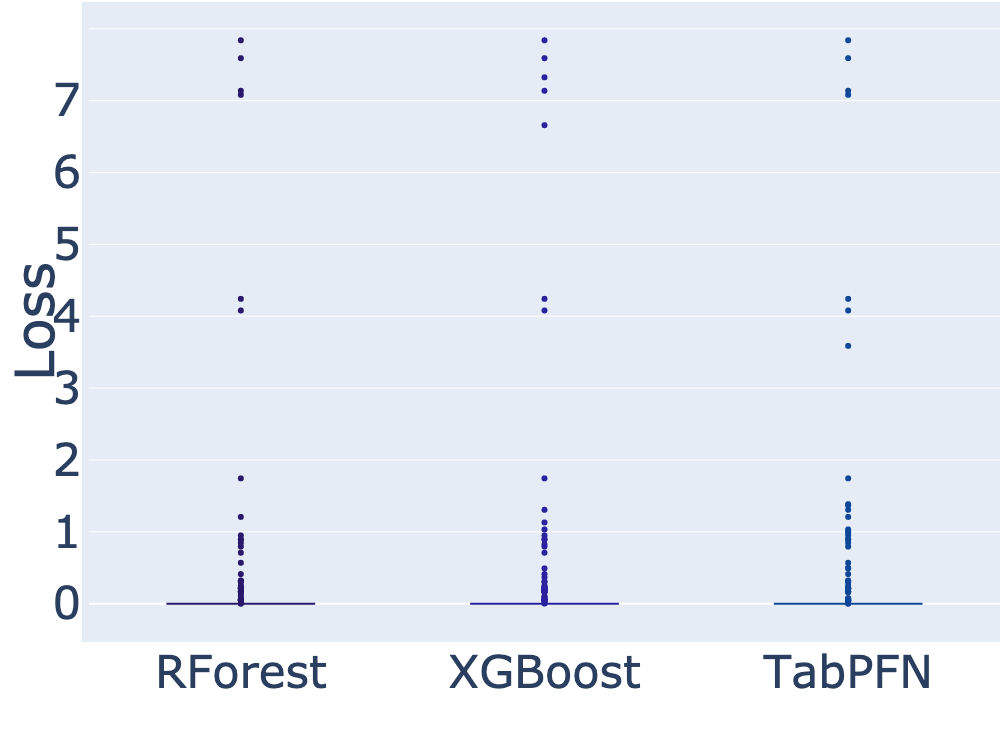}
\caption{pairwise classification}
\end{subfigure}
\caption{The loss of the AS across the ML model types for the three AS approaches (i.e., R = regression, C = classification, and PC = pairwise classification), aggregated over BBOB problems in $5D$ for runtime 2500.}
 \label{fig:boxplotMLalgo}
\end{figure*}

\begin{figure}
    \centering
    \includegraphics[width=\linewidth]{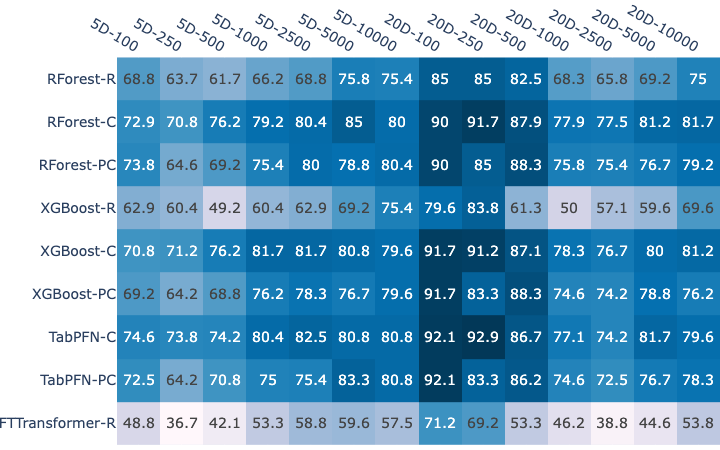}
    \caption{A heatmap depicting the percentage of problem instances where the VBS is selected as optimal by the AS, for the various problem dimensions, runtimes, ML methods, and AS approaches.}
    \label{fig:heatmapAgregated}
\end{figure}

Following the experimental protocol described in Section~\ref{sec:exp-setup}, we first affirm that AS leads to impressive performance gains compared to any standalone solver. In order to assess the performance in a principled way, we test all chosen AS approaches/ML models on 10 instances of each of the 24 BBOB problems, for different algorithm runtime values and in dimensions 5 and 10.
We compare the performance of the AS with that of different solvers in the portfolio, aggregated over all BBOB problems. We see that AS undoubtedly outperforms all solvers by a large margin, irrespective of the ML model or task. While we do not show the specific results of the AS comparison with individual solvers here due to space constraints, we make them available at~\cite{gecco2023data}.

\subsection{Comparing Algorithm Selection Approaches}
\label{sec:reg-class}

After establishing the relevance of AS itself, we compare the performance of different AS approaches (or underlying ML tasks), namely regression, classification, and pairwise classification. Contrasting the losses obtained by different AS approaches, and we do not observe any, drastic differences among them. However, regression-based AS seems to have a slightly worse performance than classification and pairwise classification. This holds true across all different ML models, all seven different runtimes, and in both dimensions~\cite{gecco2023data}.

To intuitively quantify the high performance (and thus reliability) of all considered approaches, we record the number of times that each solver from the portfolio is the VBS and the number of times each solver is selected by the AS based on regression, classification, and pairwise classification.
We observe that all AS approaches very tightly follow the overall distribution of VBSs. There are no significant differences across different runtimes for any of the ML tasks. This is consistent for other ML models as well. The results and visualizations that support these conclusions are available at~\cite{gecco2023data}. 

\subsection{Comparing Machine Learning Models}
\label{sec:ml-models}

We further investigate how different ML models compare for a fixed AS approach, across all runtimes and dimensions.
To achieve this, we contrast the losses obtained by different ML models, as shown in Figure~\ref{fig:boxplotMLalgo} for a runtime of 2500 in $5D$. 
Similarly, as in Section~\ref{sec:reg-class}, we conclude that the regression-based AS approach results in slightly larger losses compared to classification and pairwise classification, but the difference is not substantial enough to draw definite conclusions. 
Among the regression models, the RF model outperforms XGBoost. 
This can be attributed to the ease of tuning RF compared to XGBoost. 
Implementing a different hyperparameter tuning strategy, such as Bayesian Optimization, would likely further improve the performance of XGBoost and bring it closer to that of RF.

The FT-Transformer regression model shows the largest loss compared to the other two regression-based models, which is in line with our expectations. 
As highlighted in ~\cite{gorishniy2021revisiting}, tree-based models outperform deep learning models, such as the FT-Transformer, on tabular problems.
This may be partly due to the fact that DL models usually need more training data than tree-based models. 
In this study, with only 240 data examples, the FT-Transformer might not have sufficient training data to generalize effectively. 
Although the performance of the FT-Transformer could potentially be improved with a different tuning strategy, we still anticipate that the tree-based models will outperform the FT-Transformer in our AS task.
On the other hand, when it comes to classification and pairwise classification, there are no significant differences between the ML models. This remains consistent across all runtimes and dimensions, as evidenced by the results available in~\cite{gecco2023data}.

Finally, we summarize the results of the AS performance across different ML models, AS approaches, runtimes, and dimensions in Figure~\ref{fig:heatmapAgregated}.  
Here, instead of focusing on the loss as a quality indicator of the AS performance, we measure the percentage of problem instances where the VBS was selected as optimal by the AS. 
Our previous observations still hold true, i.e., regression-based AS performs worse than the other two approaches, especially in the case of FT-Transformer models (which is now clearly visible in Figure~\ref{fig:heatmapAgregated}), while classification and pairwise classification exhibit comparable performance with a slight edge for classification. 
Notably, TabPFN performs similarly to tree-based methods, or even slightly better, which has also been observed in~\cite{tabpfn}. We point out that, for small runtimes (100 and 250) in $20D$, exceptionally good performance of (almost) all techniques might be due to the small size of algorithm portfolios for respective runtimes.
With less algorithms in the portfolio, the AS problem gets easier to tackle, which is evident in Figure~\ref{fig:heatmapAgregated} from the larger percentage of correctly selected VBS.

\section{Discussion and conclusions}
\label{sec:discussion}

We have confirmed in this work that PIAS bears considerable potential for numerical black-box optimization, and that ELA-based approaches are a sensible candidate to achieve impressive performance gains. 
We have also shown that the choice of the ML model (given that it had already been shown to perform well on tabular data in general) and the AS approach (regression, classification, pairwise classification) is of minor importance. However, this does not suffice to benefit from PIAS in practice, as real-world use cases remain very costly to both optimize and extract information from.
We thus face two key challenges in applying these methods to the black-box optimization domain: (1) the cost for feature extraction and (2) warm-starting the chosen solvers. 

Since our primary objective in this work has been the comparison of the different ML techniques, we have ignored challenge (1) by assuming that ELA features are available at no cost. This is of course not true in practice, where the solvers need to balance the trade-off between sampling for ELA computation and the budget that remains for the actual optimization phase. 
The first attempts  aimed at integrating the ELA computation into the optimization process (\cite{AnjaPPSN2022} and references therein) encourage us to believe that ELA features can be valuable even when extracted from search trajectories rather than from a dedicated sampling phase that precedes the optimization, but also demonstrate much 
room for improvement. Regarding the warm-starting aspect, item (2) in our list above, we observe a surprising ignorance in the community, and this in works on per-instance algorithm selection, per-instance algorithm configuration, as well as in studies deliberately focusing on switching from one solver to another. We believe that neglecting the warm-starting aspect is wasteful. First proposals for warm-starting techniques have been made in~\cite{MohammadiRT15EgoCmaes} and~\cite{DynASDominik}. The latter was considered in the \emph{per-run} approach suggested in~\cite{AnjaPPSN2022}, but the small number of algorithms for which dedicated warm-starting techniques have been designed suggests that this can only be seen as the first starting point.
We suspect both open questions (1) and (2) are of utmost importance for the wider adoption of PIAS and per-instance algorithm configuration techniques in practice, both in academic and industrial use cases.

\begin{acks} 
The authors acknowledge the support of the Slovenian Research Agency through program grant No. P2-0103 and P2-0098, project grants N2-0239 and J2-4460, a young researcher grant (PR-09773) to AK, and a bilateral project between Slovenia and France grant No. BI-FR/23-24-PROTEUS-001 (PR-12040), as well as the EC through grant No. 952215 (TAILOR). Our work is also supported by ANR-22-ERCS-0003-01 project VARIATION.
\end{acks}

\bibliographystyle{ACM-Reference-Format}
\bibliography{references} 

\end{document}